%% file: conference.tex
\documentclass[conference]{IEEEtran}
\IEEEoverridecommandlockouts
\usepackage{cite}
\usepackage{amsmath,amssymb,amsfonts}
\usepackage{algorithmic}
\usepackage{graphicx}
\usepackage{textcomp}
\usepackage{xcolor}

\usepackage{url}
\usepackage{booktabs}
\usepackage{adjustbox}      
\usepackage{xcolor, colortbl}
\usepackage{pgf}            
\usepackage{comment}

    \makeatletter
    \newcommand{\linebreakand}{%
      \end{@IEEEauthorhalign}
      \hfill\mbox{}\par
      \mbox{}\hfill\begin{@IEEEauthorhalign}
    }
    \makeatother

\def\BibTeX{{\rm B\kern-.05em{\sc i\kern-.025em b}\kern-.08em
    T\kern-.1667em\lower.7ex\hbox{E}\kern-.125emX}}
\begin{document}

\title{A Cross-Cultural Comparison of LLM-based Public Opinion Simulation: Evaluating Chinese and U.S. Models on Diverse Societies} 


\author{\IEEEauthorblockN{Weihong Qi}
\IEEEauthorblockA{wq3@iu.edu \\ Indiana University Bloomington \\ Bloomington, IN, USA
}
\and
\IEEEauthorblockN{Fan Huang}
\IEEEauthorblockA{fanhuan@iu.edu \\
Indiana University Bloomington \\
Bloomington, IN, USA}

\linebreakand

\IEEEauthorblockN{Jisun An}
\IEEEauthorblockA{jisunan@iu.edu \\
Indiana University Bloomington \\
Bloomington, IN, USA}
\and
\IEEEauthorblockN{Haewoon Kwak}
\IEEEauthorblockA{hwkwak@iu.edu \\
Indiana University Bloomington \\
Bloomington, IN, USA}
}

\maketitle

\begin{abstract}
This study evaluates the ability of DeepSeek, an open-source large language model (LLM), to simulate public opinions in comparison to LLMs developed by major tech companies. By comparing DeepSeek-R1 and DeepSeek-V3 with Qwen2.5, GPT-4o, and Llama-3.3 and utilizing survey data from the American National Election Studies (ANES) and the Zuobiao dataset of China, we assess these models' capacity to predict public opinions on social issues in both China and the United States, highlighting their comparative capabilities between countries. Our findings indicate that DeepSeek-V3 performs best in simulating U.S. opinions on the abortion issue compared to other topics such as climate change, gun control, immigration, and services for same-sex couples, primarily because it more accurately simulates responses when provided with Democratic or liberal personas. For Chinese samples, DeepSeek-V3 performs best in simulating opinions on foreign aid and individualism but shows limitations in modeling views on capitalism, particularly failing to capture the stances of low-income and non-college-educated individuals. It does not exhibit significant differences from other models in simulating opinions on traditionalism and the free market. Further analysis reveals that all LLMs exhibit the tendency to overgeneralize a single perspective within demographic groups, often defaulting to consistent responses within groups. These findings highlight the need to mitigate cultural and demographic biases in LLM-driven public opinion modeling, calling for approaches such as more inclusive training methodologies.
\end{abstract}

\begin{IEEEkeywords}
large language model, cultural comparison, public opinions
\end{IEEEkeywords}

\input{content/1_intro}

\input{content/2_literature}

\input{content/3_method}

\input{content/4_results}

\input{content/5_conclusion}

\input{conference.bbl}
\input{content/appendix}

\end{document}

%% file: content/1_intro.tex
\section{Introduction}


The release of DeepSeek, an open-source model developed in China, marks a significant milestone in the development of large language models (LLMs), emerging as a strong competitor following the launch of ChatGPT in November 2022~\cite{liu2024deepseek}. 
Its strong performance invites broader evaluations beyond benchmark scores~\cite{bird2020fairlearn, sheng2024fairness}, particularly regarding fairness and cultural representation, as the model is expected to see widespread use due to its competitive cost. 
Recent studies have examined cultural biases in dominant models, finding that they often align more closely with English-speaking or Western-centric values~\cite{tao2024cultural, qi2024representation}.

The emergence of a competitive non-Western LLM therefore presents a critical test case: Are cultural biases an inevitable byproduct of current LLM development practices, or can a model built within a different cultural ecosystem offer a more balanced perspective?

This study addresses that question by investigating whether DeepSeek can more accurately capture public opinions across the United States and China compared to prominent U.S. and Chinese counterparts, including GPT-4o~\cite{achiam2023gpt}, Llama-3.3-70B-Instruct-Turbo~\cite{dubey2024llama}, and Qwen-2.5-72B-Instruct-Turbo~\cite{yang2024qwen2}.
Using survey data from the American National Election Studies (ANES)~\cite{ANES2020} and China's Zuobiao dataset~\cite{pan2018china}, we evaluate how these models simulate opinions on controversial social issues in both nations.

Specifically, we focus on the following research questions:
\begin{itemize}
    \item \textbf{RQ1.} Does an LLM's cultural origin create a ``home-field advantage'' in simulating public opinion within its native cultural context, or does it introduce distinct cultural biases when modeling opinions cross-culturally?
    \item \textbf{RQ2.} How do these cultural and demographic biases manifest in LLM simulations? Specifically, do models systematically overgeneralize or underrepresent certain societal groups in culturally-patterned ways?
\end{itemize}


In particular, we provide a prompt that simulates each survey participant by specifying their demographic attributes available in the survey, including ethnicity, gender, age, political leaning, income, and education level. We then pose survey questions to the LLMs, and the responses are treated as the LLM's simulated answers for that individual. Finally, we aggregate and evaluate the performance for each demographic group and LLM.

Our findings indicate that an LLM's cultural origin does not confer a straightforward `home-field advantage' in simulating public opinion.  DeepSeek, despite its development in China, fails to demonstrate a consistent or significant advantage in modeling Chinese societal views compared to its U.S.-based counterparts. Instead, our cross-cultural analysis reveals that all evaluated LLMs, regardless of origin, exhibit significant cultural and demographic biases. These models struggle to capture the nuanced perspectives within specific societal groups, often defaulting to stereotyped or overgeneralized responses.

This tendency is highlighted by DeepSeek-V3's mixed performance. While it more accurately simulates the opinions of U.S. Democrats and liberals on the issue of abortion, it simultaneously fails to represent their Republican and conservative counterparts. Similarly, in the Chinese context, most models struggle to model the views of low-income and non-college-educated individuals on capitalism. This flaw is exemplified in the extreme by Qwen2.5; although it appeared to outperform other models on this topic, its high accuracy was achieved by defaulting to a single, uniform response for these demographic groups, mistaking overgeneralization for accurate representation. These results suggest the core issue is not a simple East-West divide but a deeper, more systemic failure of current LLMs to achieve genuine cultural and demographic sensitivity.

%% file: content/2_literature.tex
\section{Related Work}

Prior to the release of DeepSeek, extensive research has been conducted on the fairness and inclusion of LLMs. For example, \cite{bird2020fairlearn, madaio2022assessing, atwood2024inducing} develop metrics and tools to evaluate AI fairness. Other studies focusing on the representation of specific social groups have identified biases related to gender~\cite{wan2023kelly}, culture~\cite{naous2023having}, race~\cite{haim2024s}, political leanings~\cite{feng2023pretraining}, political affiliations and institutions~\cite{qi2024representation}. However, existing research has two key limitations. First, although some studies have included LLMs developed by teams from non-U.S. cultural backgrounds~\cite{naous2023having, ye2024justice}, the primary focus remains on American AI models, with limited attention to potential biases stemming from differences in the training corpus and the development teams' cultural backgrounds. Second, while these studies have identified various biases~\cite{cao2023assessing, wan2023kelly, tao2024cultural}, they often fall short of uncovering the underlying sources of these biases or proposing potential solutions. The comparable performance of DeepSeek to American models offers an opportunity to further investigate these issues. Our study aims to bridge these gaps in the literature.

Representative bias can stem from various sources, one of which is how developers sample populations during the training data collection process~\cite{suresh2019framework}. Non-representative samples lead to demonstrable biases against certain groups. For example, \cite{caton2024fairness} finds that a lack of geographical diversity in datasets, such as ImageNet, results in biases favoring Western cultures. Similarly, since English-language corpora constitute the majority of training data for most popular LLMs~\cite{achiam2023gpt, touvron2023llama, taubenfeld2024systematic}, these models may exhibit biases favoring English-speaking populations. Moreover, models trained within specific cultural contexts can encode and perpetuate harmful human biases~\cite{arseniev2022machine}. Several studies have identified cultural biases and a lack of cultural understanding in LLMs~\cite{tao2024cultural, kharchenko2024well}. Our research extends these investigations by examining the newly released DeepSeek model and its simulations of public opinions. This is particularly significant as DeepSeek, developed by a team in China, offers a contrasting cultural and commercial background compared to popular LLMs like the GPT and LLaMA series. This comparison allows us to explore whether cultural differences in development teams contribute to variations in opinion simulation.

%% file: content/3_method.tex
\section{Method}

\input{Fig_tables/flow}


\subsection{Datasets}

We utilize the 2020 American National Election Studies (ANES) and the Zuobiao Dataset of China~\cite{pan2018china} as benchmarks of human ground truth to evaluate LLMs in public opinion simulation. ANES data series are academically conducted national surveys of U.S. citizens, administered before and after presidential election cycles. These surveys follow a longstanding tradition, with data collection dating back to 1948. The survey data includes multiple questions on essential issues, such as abortion, climate change, and gun control, in the election cycle. The 2020 ANES dataset consists of responses from 2,457 individuals. The Zuobiao dataset, developed by \cite{pan2018china}, is based on online surveys designed to capture the ideological spectrum of Chinese citizens across various dimensions, such as capitalism, foreign aid, traditionalism, individualism, and the free market. This dataset contains opinions and stances from over 470,000 individuals. To ensure comparability between the two datasets, we randomly sample 2,000 individuals from the Zuobiao dataset. Both datasets also contain individual-level information, such as demographics and socioeconomic status, which is essential in public opinion research~\cite{erikson2019american, mccombs2020setting}. 


We use survey questions from both datasets that are considered representative of public opinion at the time of data collection~\cite{ANES2020, pan2018china} for our simulations and evaluations. From the ANES dataset, we include questions on abortion, climate change, gun control, immigration, and services for same-sex couples. From the Zuobiao dataset, we include questions on capitalism, foreign aid, traditionalism, individualism, and the free market. Detailed descriptions of these 10 questions are provided in the Appendix A.

For the individual-level simulation using LLMs, we extract individual attributes from both datasets. From the ANES dataset, we include race, gender, age, educational level, income, religious beliefs, ideology, and partisanship. From the Zuobiao dataset, we include gender, age, educational level, and income. The feature selection is consistent with the political science research using these datasets~\cite{pan2018china, bisbee2024synthetic}, which also considers them to be important factors influencing public opinions.  The discrepancies in the selected features between the American and Chinese samples result from adaptations to the social and cultural contexts of the two countries. Certain attributes, such as race and religious beliefs, are more salient and exhibit greater variability within the U.S. population than in China. For example, the majority of the Chinese population belongs to the Han ethnic group (91.11\% as of 2021)~\cite{china_population2023}. Only 10\% of Chinese adults identified with any religious group in the 2018 Chinese General Social Survey (CGSS)~\cite{pew_research_china_religion_2023}. Figure~\ref{fig: flow_chart} summarizes our evaluation process for the LLMs in public opinion simulation.

\subsection{Experimental Design}
We incorporate individual information into the LLM prompts to reflect demographic variations within populations. Due to space constraints, we present the English prompt for the ANES dataset here, while the survey questions and corresponding prompts for the Chinese dataset are provided in the Appendix section. An example of the designed {\tt prompt} for public opinion simulation is as follows:

\begin{center}
\begin{minipage}{.4\textwidth} 
    {\tt System:} Imagine you are a(n) \{{\tt country}\} citizen. \\
    {\tt Prompt:}  You are a \{{\tt race}\} individual and identify as \{{\tt gender}\}. You are \{{\tt age}\}, have \{{\tt highest degree}\}, and your annual income is \{{\tt income}\}. Your religious beliefs align with \{{\tt religious belief}\}. Politically, you identify as \{{\tt ideology}\} and \{{\tt party}\} in party alignment. What is your opinion to the question: \{{\tt Question}\}. Only respond one of the three answer: \{{\tt Options}\}.
\end{minipage}
\end{center}



We then compute the simulation accuracies using the following method:

\begin{equation}
Accuracy = \frac{\sum_i^{N} \mathbb{I}_i(M = H)}{N}
\end{equation}

In this formula, $M$ represent the simulated opinion by the LLM models, $H$ represents the ground truth opinion of humans, and $\mathbb{I}_i$ is an indicator function that equals 1 if the two values are considered equivalent, and 0 otherwise. $N$ denotes the total number of observations. 

For subgroup accuracy, we calculate the accuracy based solely on the specified subpopulation:

\begin{equation}
Accuracy_G = \frac{\sum_i^{N_G} \mathbb{I}_i(M_G = H_G)}{N_G}
\end{equation}

In Equation (2), $M_G$, $H_G$, and $N_G$ also denote the simulated opinions by the LLMs, the ground truth human opinions, and the total number of individuals, respectively, but restricted to the specified subgroup $G$.

By comparing the simulation accuracies of different models on questions from diverse social and cultural contexts, we evaluate whether DeepSeek can simulate public opinion better than other LLMs. Furthermore, when significant discrepancies arise between human opinions and those simulated by DeepSeek---or between DeepSeek and other models---we break down the accuracies by demographic groups to investigate the sources of these differences. This analysis helps identify potential biases toward specific demographic groups within the LLMs.

\subsection{Models}

Our study involves experiments with five LLMs:  DeepSeek-R1, DeepSeek-V3, Qwen2.5-72B-Instruct-Turbo, Llama-3.3-70B-Instruct-Turbo, GPT-4o. For simplicity, we refer to Qwen2.5-72B-Instruct-Turbo as Qwen2.5 and Llama-3.3-70B-Instruct-Turbo as Llama-3.3. Among these models, the two DeepSeek models are the primary focus of this study. They represent a significant competitor to OpenAI's GPT series and are open-source. As comparisons, Qwen2.5 provides a perspective as another open-source LLM developed in China. GPT-4o and Llama-3.3 represent models developed by U.S.-based tech companies. 

%% file: Fig_tables/flow.tex
\begin{figure}[h]
    \centering
    \includegraphics[width= \linewidth]{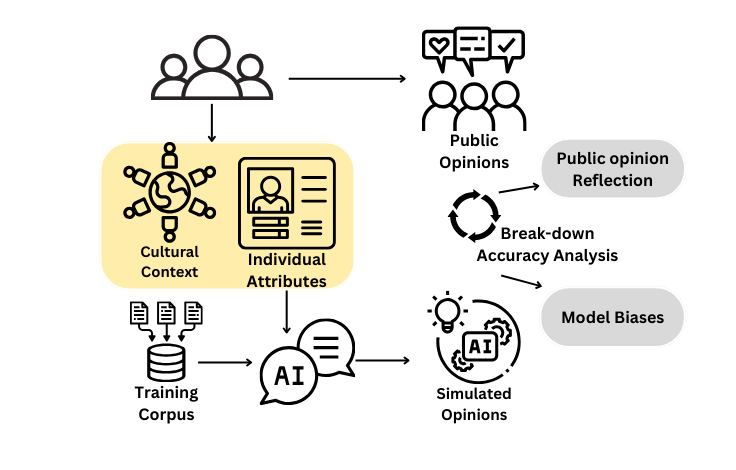}
    \caption{Overview of the Evaluation Process for LLMs in Public Opinion Simulation}
    \label{fig: flow_chart}
\end{figure}

%% file: content/4_results.tex
\section{Results}

\subsection{Overall Performance}

\input{Fig_tables/all_accuracy}

\input{Tables/anes_acc}

\input{Tables/china_acc}

We begin by examining the simulation accuracy of all five models. Notably, DeepSeek-R1 shows significantly lower accuracy across all topics, primarily due to a high proportion of invalid responses (averaging 57\% and reaching as high as 99.6\%), where the model refuses to give the requested answers. Detailed results are provided in Appendix B. The model's long-form reasoning style, which often results in responses exceeding the output window. The potential in-built censorship mechanism of LLMs, particularly in responses to questions related to China, appears to be the main cause of these invalid outputs. These findings suggest that reasoning-focused models like DeepSeek-R1 require further prompt adaptation for effective use in public opinion simulations and may be more prone to content filtering when addressing sensitive topics. Due to the high share of invalid responses of DeepSeek-R1, the remainder of this section focuses on the other four models for further analysis.

Figure~\ref{fig: all_accuracy} presents the overall simulation accuracies of GPT-4o, DeepSeek-V3, Qwen2.5, and Llama-3.3 on five public opinion questions from American and Chinese samples, respectively. Specifically, DeepSeek does not exhibit significant differences with other models in most of the topics, with slight advantages of 0.01 in accuracy in simulating opinions on foreign aid and individualism in the Zuobiao dataset. However, there are two notable exceptions. On the abortion issue in the ANES dataset, DeepSeek achieves a significantly higher accuracy of 0.53, compared to a maximum of 0.46 among the other models. Conversely, it underperforms compared to {\tt Qwen2.5} in simulating opinions on capitalism in the Zuobiao dataset, with an accuracy of 0.36 versus 0.54.


\subsection{Which Groups Are Accurately Simulated, and Which Are Not?}

As shown in Figure~\ref{fig: all_accuracy}, DeepSeek-V3 demonstrates a significant advantage in simulating opinions on the abortion issue with American samples but performs poorly on the capitalism question with Chinese samples, particularly in comparison to Qwen2.5. 
Why does DeepSeek-V3 perform well on some issues but poorly on others? 
To further investigate the sources of these discrepancies and identify potential biases in the LLMs, we conduct a group-level breakdown of simulation accuracy across demographic features.

\input{Fig_tables/abortion_diff}

\input{Fig_tables/capitalism_diff}



As shown in Figure~\ref{fig: abortion_diff}, the largest differences between DeepSeek and GPT-4o in public opinion simulation on the abortion issue appear among subgroups defined by partisanship and ideology. Although these two features are highly correlated (Cramer's V= 0.54), we keep them separate in the analysis because they are treated as distinct variables in the original survey data.

Table~\ref{tab: abortion} presents the simulation accuracies by party affiliation and ideology for the abortion issue using American samples, where all models exhibiting their lowest accuracy on this topic. The breakdown accuracy with other attributes is reported in Table~\ref{tab: app_anes} in the Appendix. The accurate simulation performance of DeepSeek-V3 is primarily driven by its accurate representation of opinions from Democratic and liberal respondents. As shown in the confusion matrix in Figure~\ref{fig: abortion_diff}, while other LLMs, such as GPT-4o, tend to underestimate the liberal alignment of Democrats and liberals, DeepSeek-V3 predicts their responses more accurately. 
Llama-3.3 shows a similar pattern, though with lower accuracy than DeepSeek-V3. 
However, when simulating the opinions of Republicans and conservatives, we observe the opposite trend across the models. 
DeepSeek-V3 and Llama-3.3, which perform better for Democrats and liberals, exhibit lower accuracy for Republicans and conservatives (See Table~\ref{tab: abortion}). 
In contrast, Qwen2.5 and GPT-4o achieve higher accuracy for Republicans and conservatives, but perform worse for Democrats and liberals.


For the capitalism issue in China, income and education level show the largest performance gaps among all demographic features, as shown in Table~\ref{tab: china_acc} and Table~\ref{tab:app_zuobiao} in the Appendix.



As low-income individuals make up 43.9\% of the full population, the results indicate that DeepSeek-V3's poor performance in representing Chinese public opinions on capitalism is primarily due to its low accuracy in simulating the opinions of low-income individuals and those without a college degree. A similar pattern is observed in the performance of  GPT-4o and Llama-3.3. Only Qwen2.5 demonstrates a notably accurate representation of low-income individuals and those without a college degree. This raises an important question: does Qwen2.5 provide a fairer representation of marginalized populations compared to the other models?

Contrary to what one might expect, the answer is no. Rather, a closer analysis reveals an undesirable pattern in Qwen2.5's behavior. 
The confusion matrix in Figure~\ref{fig: capitalism_diff}  shows that  Qwen2.5 returns the same response to 99.89\% and 96.15\% of all survey questions when the prompt specifies low income and no college degree, regardless of any other demographic information.  



\subsection{Evaluating the Prompts Language Sensitivity in LLM Simulations }

Prior research has shown that training data in different languages can introduce distinct biases in LLMs~\cite{kharchenko2024well}, and the use of different languages could result in significantly different outputs of LLMs~\cite{zhong2024cultural}. To explore this, we repeat the experiments with the Zuobiao dataset using prompts in simplified Chinese and compare the outcomes to those obtained using English prompts.

\input{Fig_tables/CN_acc}
Figure~\ref{fig: chinese_accuracy} presents the simulation results using Chinese prompts. Overall, Chinese prompts do not lead to consistent and significant improvements in simulation accuracy across tasks. The exceptions are that Chinese prompts enhance simulation accuracy for most LLMs on the capitalism issue and improve performance on individualism for both Qwen2.5 and GPT-4o. However, using Chinese prompts does not offer a notable advantage for DeepSeek-V3 in most tasks. These results suggest that changing the prompt language can influence model performance on certain topics, but does not lead to consistent improvements across all areas. Therefore, simply switching languages does not reliably enhance simulation accuracy across cultural contexts.

%% file: Fig_tables/all_accuracy.tex
\begin{figure*}[t]
  \includegraphics[width=0.48\linewidth]{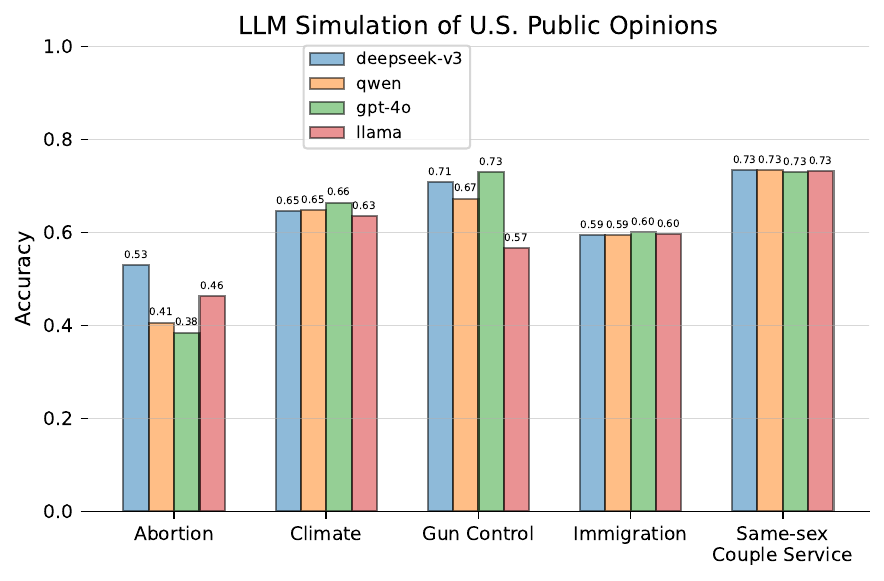} \hfill
  \includegraphics[width=0.48\linewidth]{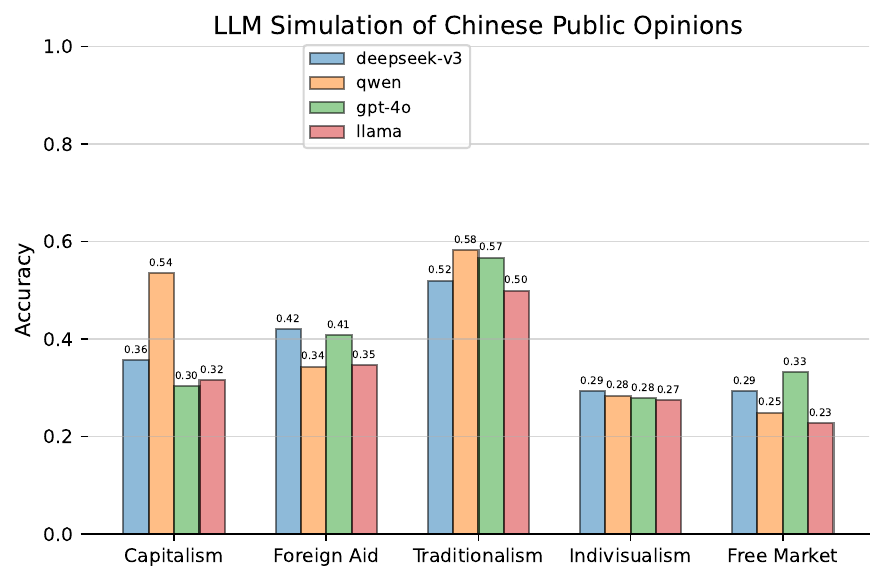}
  \caption {Simulation results for US and Chinese samples in public opinions}
  \label{fig: all_accuracy}
\end{figure*}

%% file: Tables/anes_acc.tex
\begin{table*}
\centering
\caption{Simulation Accuracy of Public Opinions on Abortion by Party Affiliation and Ideology with U.S. Samples} 
\adjustbox{max width=\linewidth}{
\begin{tabular}{l | c c c | c c c}
\toprule[1.1pt]
 & \multicolumn{3}{c |}{Party} & \multicolumn{3}{c}{Ideology} \\
\textbf{Model}   & Democrat & Republican & Independent & Liberal & Conservative & Moderate \\
\midrule
DeepSeek-V3 & \cellcolor{blue!15}0.6510 & \cellcolor{blue!8}0.3956 & \cellcolor{blue!12}0.5286 & \cellcolor{blue!20}0.7301 & \cellcolor{blue!10}0.4207 & \cellcolor{blue!5}0.3614 \\
Qwen2.5     & \cellcolor{blue!3}0.3268  & \cellcolor{blue!11}0.5267 & \cellcolor{blue!6}0.3711  & \cellcolor{blue!2}0.3132  & \cellcolor{blue!13}0.5350 & \cellcolor{blue!4}0.3560 \\
GPT-4o      & \cellcolor{blue!1}0.2232  & \cellcolor{blue!14}0.5983 & \cellcolor{blue!7}0.3488  & \cellcolor{blue!0}0.1984  & \cellcolor{blue!15}0.5989 & \cellcolor{blue!4}0.3560 \\
Llama-3.3   & \cellcolor{blue!14}0.5930 & \cellcolor{blue!4}0.3263  & \cellcolor{blue!9}0.4537  & \cellcolor{blue!18}0.6483 & \cellcolor{blue!5}0.3326  & \cellcolor{blue!4}0.3560 \\

\midrule
Group Size & 914 & 809 & 734 & 978 & 559 & 920\\

\bottomrule[1.1pt]
\end{tabular}}
\label{tab: abortion}
\end{table*}

%% file: Tables/china_acc.tex
\begin{table*}
\centering
\caption{Simulation Accuracy of Public Opinions on Capitalism by Income and Education Level with Chinese Samples} 
\adjustbox{max width= .95\linewidth}{
\begin{tabular}{l | c c c | c c}
\toprule[1.1pt]
 & \multicolumn{3}{c |}{Income} & \multicolumn{2}{c}{Education} \\
\textbf{Model} & Low Income & Medium Income & High Income & No College Degree & College Degree \\

\midrule
DeepSeek-V3 & \cellcolor{blue!2}0.1390 & \cellcolor{blue!13}0.5377 & \cellcolor{blue!12}0.5028 & \cellcolor{blue!3}0.2396 & \cellcolor{blue!6}0.3309 \\
Qwen2.5    &  \cellcolor{blue!15}0.5547 & \cellcolor{blue!16}0.5662 & \cellcolor{blue!9}0.4176 & \cellcolor{blue!15}0.5503 & \cellcolor{blue!13}0.5319 \\
GPT-4o   & \cellcolor{blue!1}0.1241 & \cellcolor{blue!9}0.4156 & \cellcolor{blue!12}0.5028 & \cellcolor{blue!2}0.1598 & \cellcolor{blue!6}0.3321 \\
Llama-3.3   & \cellcolor{blue!1}0.1116 & \cellcolor{blue!16}0.5688 & \cellcolor{blue!4}0.2699 & \cellcolor{blue!3}0.2396 & \cellcolor{blue!6}0.3309 \\

\midrule 

Group Size & 878 & 770 & 352 & 338 & 1662 \\

\bottomrule[1.1pt]
\end{tabular}}
\label{tab: china_acc}
\end{table*}

%% file: Fig_tables/abortion_diff.tex
\begin{figure}
    \centering
    \includegraphics[width= .9 \linewidth]{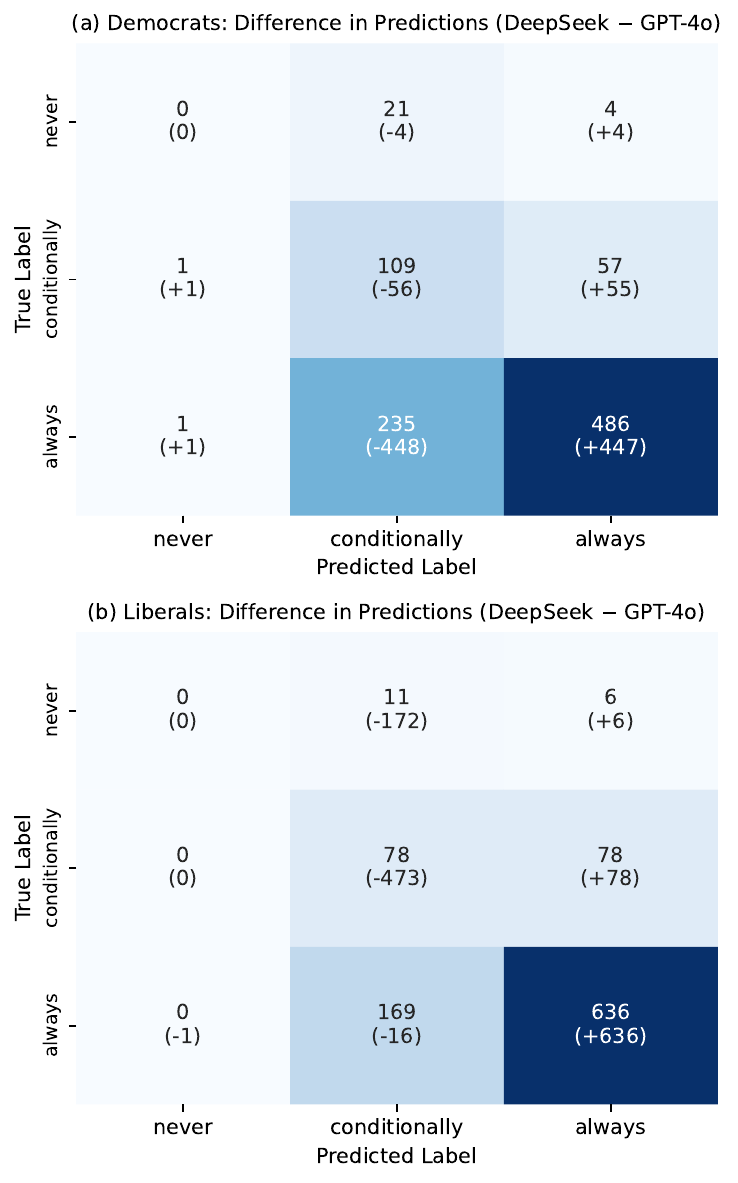}
    \caption{Confusion Matrices for Public Opinion Simulation Differences Using U.S. Data with DeepSeek and GPT-4o. The values are DeepSeek predictions, and the values in parentheses represent the difference: DeepSeek - GPT-4o. The question on Abortion issue is: Should abortion permitted by law? Only respond one of the three answers: Should never be permitted, should be permitted conditionally, should always permitted.}
    \label{fig: abortion_diff}
\end{figure}

%% file: Fig_tables/capitalism_diff.tex
\begin{figure}
    \centering
    \includegraphics[width= .9 \linewidth]{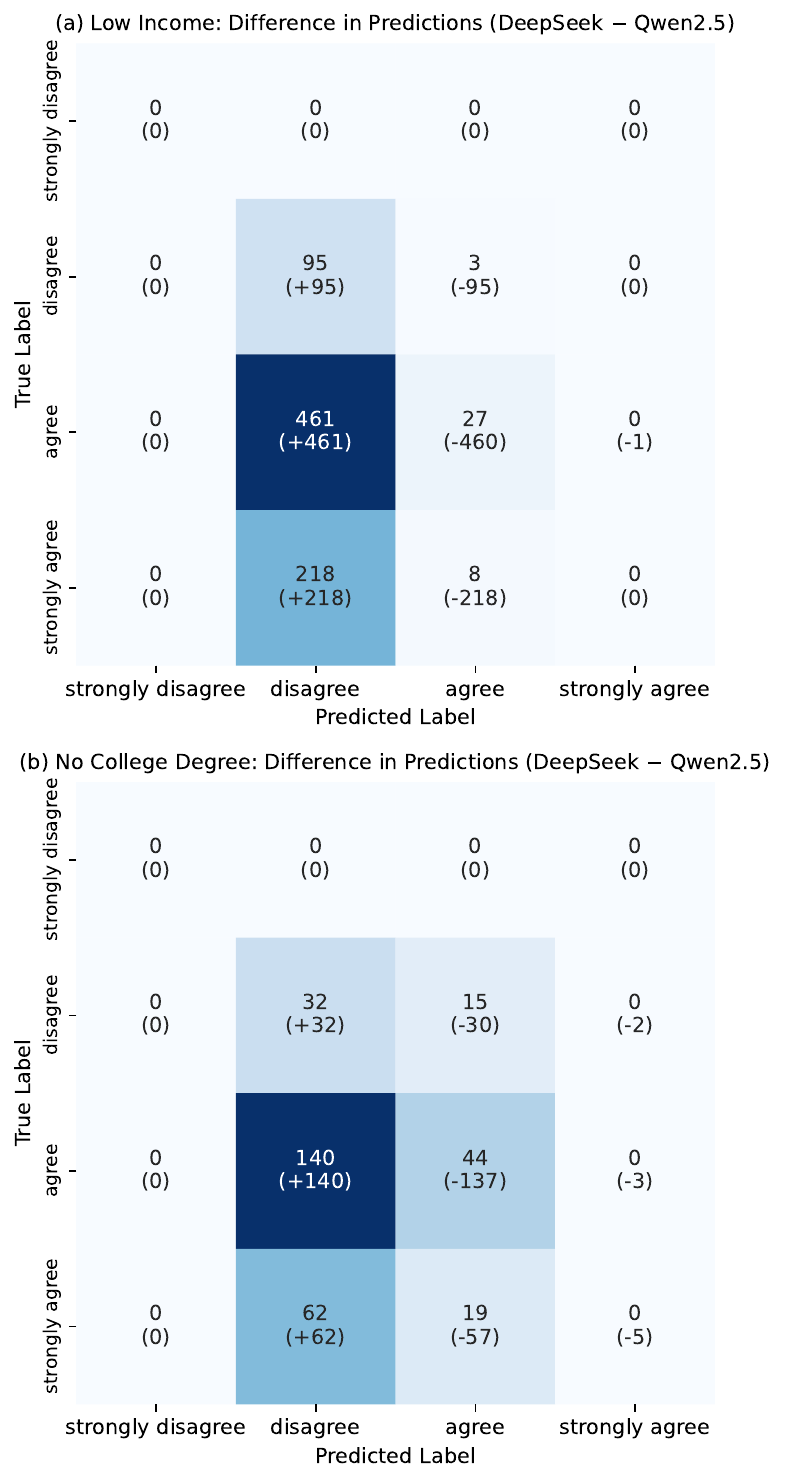}
    \caption{Confusion Matrices for Public Opinion Simulation Differences Using Zuobiao Data with DeepSeek and Qwen2.5. The values are DeepSeek predictions, and the values in parentheses represent the difference: DeepSeek - Qwen2.5. The question on Capitalism is: Do you agree or disagree the statement: A rich person deserves better medical services. Respond one of the four answers: (1) strongly disagree, (2) disagree, (3) agree, or (4) strongly agree.}
    \label{fig: capitalism_diff}
\end{figure}

%% file: Fig_tables/CN_acc.tex
\begin{figure*}[t]
  \includegraphics[width=0.45\linewidth]{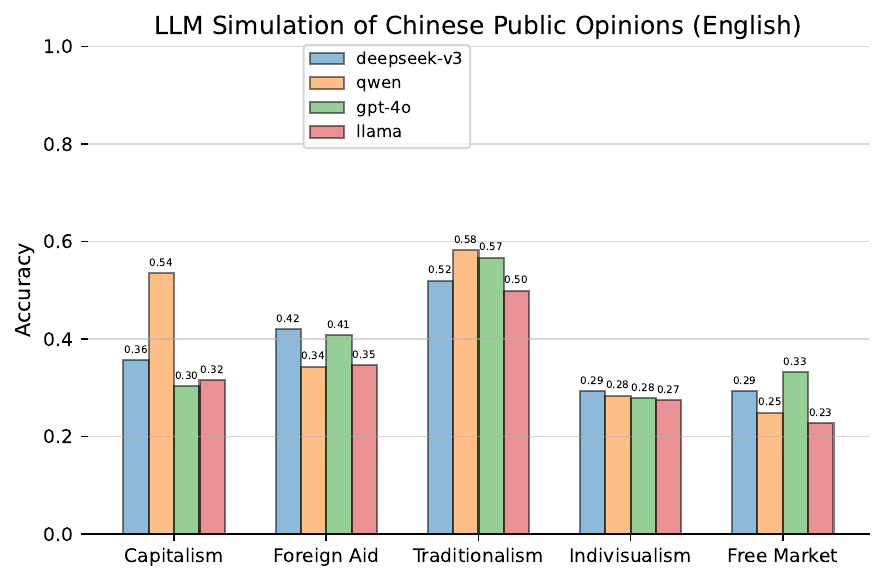} \hfill
  \includegraphics[width=0.45\linewidth]{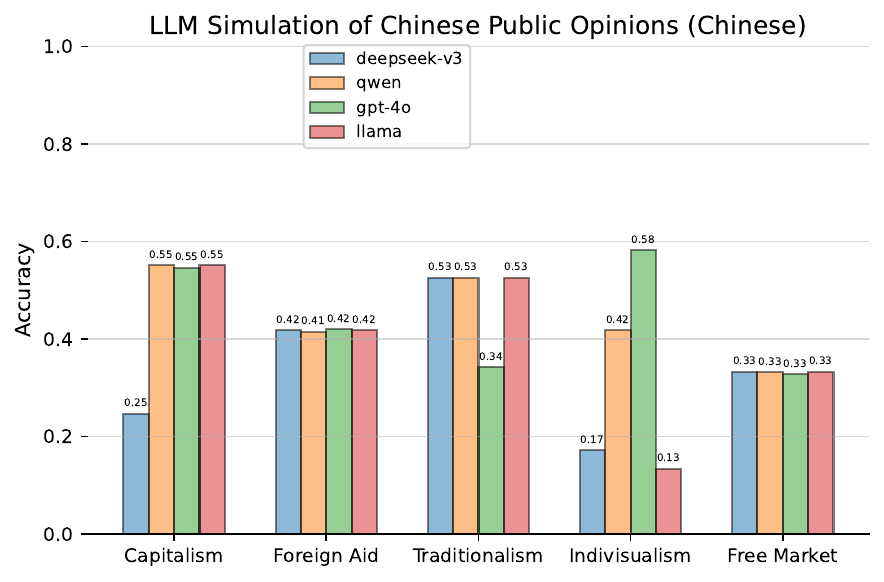}
  \caption {Simulation results for Chinese samples in public opinions with English and Chinese Prompts}
  \label{fig: chinese_accuracy}
\end{figure*}

%% file: content/5_conclusion.tex
\section{Conclusions and Discussions}
In this study, we investigate the capability of DeepSeek, an open-source LLM, to simulate public opinions compared to other prominent LLMs, including  GPT-4o, Qwen2.5, and  Llama-3.3. Using public opinion survey data from the American National Election Studies (ANES) and the Zuobiao dataset of China, we evaluate the models' accuracy in simulating responses across key social and political issues. Our analysis highlights both strengths and limitations of DeepSeek, offering insights into possible model biases and cross-cultural representation.

Our results show that DeepSeek demonstrates accurate simulation performance on certain issues, but poor performance with others. Specifically, DeepSeek achieves the highest accuracy in simulating public opinions on abortion among U.S. respondents, outperforming other prominent LLMs by effectively capturing the perspectives of Democratic and liberal respondents. However, it performs poorly in simulating Republican and conservative opinions, where GPT-4o and  Qwen2.5 show better alignment. This highlights that DeepSeek's advantages in some demographic segments are accompanied by significant limitations in others. 

In contrast, for the Chinese samples, DeepSeek performs best on questions related to foreign aid and individualism. Yet, it exhibits notable weaknesses in simulating other opinions, particularly among low-income individuals and those without a college degree. Our breakdown analysis of model performance by income and education reveals that DeepSeek and other LLMs fail to capture the diversity of opinions within these demographic groups. Although  Qwen2.5 appears to perform better, the confusion matrix analysis shows that it, too, defaults to a uniform response pattern, indicating self-reinforcing biases rather than a genuine representation of demographic differences. Our results also indicate that simply switching the prompt language to Chinese does not make a consistent improvement in simulation accuracy for the Chinese samples.

These findings highlight that demographic underrepresentation and cultural biases remain critical challenges for LLMs, regardless of their geographic origin or training methodologies. While DeepSeek performs comparably to or better than U.S.-developed models on certain issues, it does not demonstrate a distinct advantage in simulating Chinese public opinion. Instead, both Chinese- and U.S.-developed models show biases driven by overgeneralization and limited sensitivity to demographic diversity. Overall, our study highlights the importance of addressing demographic and cultural biases in LLMs to improve public opinion simulations and representations. As LLMs continue to play a central role in human-AI interactions, future research should focus on refining training datasets and modeling techniques to mitigate biases and enhance their ability to represent diverse global perspectives accurately.

%% file: content/appendix.tex
\section*{Appendix}

\subsection{Survey Questions on Public Opinions}
\label{sec: survey_q}
\textbf{Public Opinion Questions for the United States Sample.} The following questions from the ANES dataset are used in our experiments, including abortion, climate change, gun control, immigration, and services for same-sex couples:
\begin{itemize}
    \item Immigration: What is your opinion to the question: Favor or oppose returning unauthorized immigrants to native country? Only respond one of the three answer: Favor, Oppose, Neither favor nor oppose.
    \item Gun control: Should federal government make it more difficult or easier to buy a gun? Only respond one of the three answer: More difficult, Easier, or keep these rules about the same.
    \item Abortion: Should abortion permitted by law? Only respond one of the three answers: Should never be permitted, should be permitted conditionally, should always permitted.
    \item Climate change: How much is climate change affecting severe weather/temperatures in US? Only respond the number of one of the three answers: (1) Little or not at all, (2) Moderate, (3) A lot or a great deal.
    \item Same-sex couple service: Do you think business owners who provide wedding-related services should be allowed to refuse services to same-sex couples if same-sex marriage violates their religious beliefs, or do you think business owners should be required to provide services regardless of a couple’s sexual orientation? Only respond one of the two answers: Should be allowed to refuse or Should be required to provide service.
\end{itemize}

\textbf{Public Opinion Questions for the Chinese Sample.} The following questions from the Zuobiao dataset are used in our experiments, focusing on public opinions in China, including capitalism, foreign aid, traditionalism, individualism, and the free market: 
\begin{itemize}
    \item Capitalism: Do you agree or disagree the statement: A rich person deserves better medical services. Only respond one of the four answers: (1) strongly disagree, (2) disagree, (3) agree, or (4) strongly agree.
    \item Foreign Aid:  Do you agree or disagree the statement: The state has an obligation to provide foreign aid. Only respond one of the four answers: (1) strongly disagree, (2) disagree, (3) agree, or (4) strongly agree.
    \item Traditionalism: Do you agree or disagree the statement: Traditional Chinese classics should be the basic education material for children. Only respond one of the four answers: (1) strongly disagree, (2) disagree, (3) agree, or (4) strongly agree.
    \item Individualism: Do you agree or disagree the statement: The fundamental standard to evaluate the value of a work of art is whether it is liked. Only respond one of the four answers: (1) strongly disagree, (2) disagree, (3) agree, or (4) strongly agree.
    \item Free Market: Do you agree or disagree the statement: The minimum wage should be set by the state. Only respond one of the four answers: (1) strongly disagree, (2) disagree, (3) agree, or (4) strongly agree.
\end{itemize}

\subsection{Additional Results of Public Opinion Simulations}

\label{sec: more_results}

\input{Tables/appendix_anes}

\input{Tables/appendix_zuobiao}

\input{Tables/appendix_nums_nas_anes}

\input{Tables/appendix_nums_nas_china}

DeepSeek-R1 exhibits substantially lower accuracy compared to the other models. This is primarily due to the large number of invalid responses it generates when using the same prompts applied to all models. Tables~\ref{tab: nums_na_anes} and Table~\ref{tab: nums_na_china} present the counts of invalid responses across all models evaluated in the study. DeepSeek-R1, in particular, produces a high proportion of invalid responses, with the maximum proportion being 99.6\%, which significantly reduces its overall simulation accuracy.

%% file: Tables/appendix_anes.tex
\begin{table}
\caption{Simulation Accuracy of Public Opinions on Abortion with U.S. Samples by Demographics and Socioeconomic Status} 
\centering
\adjustbox{max width=\linewidth}{
\begin{tabular}{l c c c c}
\toprule[1.1pt] 
  & \textbf{DeepSeek-V3} & \textbf{Qwen2.5} & \textbf{GPT-4o} & \textbf{Llama-3.3} \\
\midrule
Female            & \cellcolor{blue!13}0.5488 & \cellcolor{blue!9}0.4179 & \cellcolor{blue!8}0.4128 & \cellcolor{blue!10}0.4490  \\
Male              & \cellcolor{blue!12}0.5105 & \cellcolor{blue!7}0.3943 & \cellcolor{blue!5}0.3575 & \cellcolor{blue!11}0.4772 \\
\midrule
White             & \cellcolor{blue!13}0.5339 & \cellcolor{blue!9}0.4183 & \cellcolor{blue!8}0.4019 & \cellcolor{blue!11}0.4681 \\
Black             & \cellcolor{blue!13}0.5437 & \cellcolor{blue!3}0.2427 & \cellcolor{blue!2}0.2330 & \cellcolor{blue!7}0.3883  \\
Hispanic          & \cellcolor{blue!11}0.4923 & \cellcolor{blue!7}0.3949 & \cellcolor{blue!5}0.3590 & \cellcolor{blue!11}0.4615  \\
Asian             & \cellcolor{blue!13}0.5421 & \cellcolor{blue!6}0.3832 & \cellcolor{blue!4}0.2804 & \cellcolor{blue!11}0.4766 \\
Native American   & \cellcolor{blue!10}0.4773 & \cellcolor{blue!5}0.3182 & \cellcolor{blue!4}0.2955 & \cellcolor{blue!8}0.4091 \\
\midrule
18–30 years old   & \cellcolor{blue!16}0.6278 & \cellcolor{blue!10}0.4479 & \cellcolor{blue!4}0.3091 & \cellcolor{blue!15}0.5836 \\
30-45 years old   & \cellcolor{blue!14}0.5583 & \cellcolor{blue!8}0.4013 & \cellcolor{blue!5}0.3317 & \cellcolor{blue!13}0.5217 \\
45-60 years old   & \cellcolor{blue!12}0.5162 & \cellcolor{blue!8}0.4051 & \cellcolor{blue!7}0.3885 & \cellcolor{blue!10}0.4478 \\
Over 60 years old & \cellcolor{blue!11}0.4898 & \cellcolor{blue!7}0.3951 & \cellcolor{blue!9}0.4380 & \cellcolor{blue!7}0.3984 \\
\midrule
No college degree & \cellcolor{blue!9}0.4591  & \cellcolor{blue!7}0.3930 & \cellcolor{blue!9}0.4339 & \cellcolor{blue!8}0.4038 \\
College degree    & \cellcolor{blue!14}0.5668 & \cellcolor{blue!9}0.4122 & \cellcolor{blue!5}0.3588 & \cellcolor{blue!12}0.4942 \\
\midrule
Less than \$35,000        & \cellcolor{blue!10}0.4856 & \cellcolor{blue!5}0.3424 & \cellcolor{blue!7}0.3916 & \cellcolor{blue!9}0.4203 \\
\$35,000 to \$90,000      & \cellcolor{blue!12}0.5236 & \cellcolor{blue!9}0.4124 & \cellcolor{blue!8}0.3970 & \cellcolor{blue!11}0.4591 \\
Above \$90,000            & \cellcolor{blue!14}0.5576 & \cellcolor{blue!10}0.4310 & \cellcolor{blue!6}0.3702 & \cellcolor{blue!11}0.4883 \\
\bottomrule[1.1pt]
\end{tabular}}
\label{tab: app_anes}
\end{table}

%% file: Tables/appendix_zuobiao.tex
\begin{table}
\centering
\caption{Simulation Accuracy of Public Opinions on Capitalism with Chinese Samples by Demographics and Socioeconomic Status} 
\adjustbox{max width=\linewidth}{
\begin{tabular}{l c c c c}
\toprule[1.1pt] 
 & \textbf{DeepSeek-V3} & \textbf{Qwen2.5} & \textbf{GPT-4o} & \textbf{Llama-3.3} \\
\midrule
Female            & \cellcolor{blue!6}0.3119  & \cellcolor{blue!14}0.5491  & \cellcolor{blue!3}0.2331  & \cellcolor{blue!6}0.3129 \\
Male              & \cellcolor{blue!8}0.3992  & \cellcolor{blue!13}0.5215  & \cellcolor{blue!7}0.3699  & \cellcolor{blue!6}0.3180 \\
\midrule
18–30 years old   & \cellcolor{blue!4}0.2263  & \cellcolor{blue!11}0.4938  & \cellcolor{blue!4}0.2325  & \cellcolor{blue!5}0.2531 \\
30-45 years old   & \cellcolor{blue!7}0.3735  & \cellcolor{blue!14}0.5453  & \cellcolor{blue!5}0.2963  & \cellcolor{blue!6}0.3058 \\
45-60 years old   & \cellcolor{blue!9}0.4219  & \cellcolor{blue!15}0.5510  & \cellcolor{blue!7}0.3535  & \cellcolor{blue!8}0.3639 \\
\bottomrule[1.1pt]
\end{tabular}}
\label{tab:app_zuobiao}
\end{table}

%% file: Tables/appendix_nums_nas_anes.tex
\begin{table}
\centering
\caption{Number of Invalid Responses in Public Opinion Simulation using ANES Data} 
\adjustbox{max width= .8\linewidth}{
\begin{tabular}{llcc}
\hline
\textbf{Task} & \textbf{model} & \textbf{\# of Invalid Responses} & \textbf{\% of Invalid Responses} \\
\hline
  Immigration & DeepSeek-R1 & 0 & 0\\
     & DeepSeek-V3 & 0 & 0\\
     & Qwen2.5 & 0 & 0\\
     & Llama-3.3 & 0 & 0\\
     & GPT-4o & 0 & 0\\
     \midrule
     
  Gun Control & DeepSeek-R1 & 736 & 30.0\% \\
     & DeepSeek-V3 & 0 & 0 \\
     & Llama-3.3 & 3 & 0.1\% \\
     & Qwen2.5 & 0 & 0\\
     & GPT-4o & 0 & 0 \\
     \midrule
     
  Abortion & DeepSeek-R1 & 2447 & 99.6\% \\
     & DeepSeek-V3 & 0 & 0\\
     & Llama-3.3 & 0 & 0\\
    & Qwen2.5 & 7 & 0.3\% \\
    & GPT-4o & 0 & 0 \\
    \midrule

  Climate    & DeepSeek-R1 & 1560 & 63.5\% \\
     & DeepSeek-V3 & 0 & 0\\
    & Llama-3.3 & 0 & 0\\
     & GPT-4o & 226 & 9.2\% \\
     & Qwen2.5 & 0 & 0\\
     \midrule
     
  LGBT & DeepSeek-R1 & 2259 & 91.9\% \\
     & DeepSeek-V3 & 0 & 0\\
     & Llama-3.3 & 0 & 0\\
     & GPT-4o & 0 & 0\\
     & Qwen2.5 & 0 & 0 \\
\hline
\end{tabular}}
\label{tab: nums_na_anes}
\end{table}

%% file: Tables/appendix_nums_nas_china.tex
\begin{table}
\centering
\caption{Number of Invalid Responses in Public Opinion Simulation using Zuobiao Data} 
\adjustbox{max width= \linewidth}{
\begin{tabular}{llc}
\hline
\textbf{Task} & \textbf{model} & \textbf{\# of Invalid Responses} \\
\hline
Free Market & DeepSeek-R1 & 396 \\
     & DeepSeek-V3 & 0 \\
     & Llama-3.3 & 0 \\
     & Qwen2.5 & 0 \\
    & GPT-4o & 9 \\
    \midrule
    
Individualism & DeepSeek-R1 & 1397 \\
     & DeepSeek-V3 & 0 \\
     & Qwen2.5 & 0 \\
     & Llama-3.3 & 0 \\
     & GPT-4o & 3 \\
     \midrule
     
Capitalism & DeepSeek-R1 & 1618 \\
     & DeepSeek-V3 & 0 \\
     & Qwen2.5 & 0 \\
     & Llama-3.3 & 0 \\
     & GPT-4o & 0 \\
     \midrule
     
Foreign Aid & DeepSeek-R1 & 1301 \\
     & DeepSeek-V3 & 0 \\
     & Qwen2.5 & 0 \\
     & Llama-3.3 & 0 \\
     & GPT-4o & 17 \\
     \midrule
     
Traditionalism & DeepSeek-R1 & 48 \\
     & DeepSeek-V3 & 0 \\
     & Qwen2.5 & 0 \\
     & Llama-3.3 & 0 \\
     & GPT-4o & 183 \\
\hline
\end{tabular}}
\label{tab: nums_na_china}
\end{table}